\documentclass[lettersize,journal]{IEEEtran}

\usepackage{array}
\usepackage{amsmath, amssymb}
\usepackage{algorithm, algpseudocode}
\usepackage{url}
\usepackage[caption=false,font=normalsize,labelfont=sf,textfont=sf]{subfig}
\usepackage{textcomp}
\usepackage{stfloats}
\usepackage{url}
\usepackage{verbatim}
\usepackage{graphicx}
\usepackage{cite}
\usepackage{booktabs}
\usepackage{hyperref}
\usepackage{balance}
\usepackage{orcidlink}
\usepackage{amsthm}
\renewenvironment{proof}{{\bfseries \textit{ Proof:}}}{\qed}

\begin{document}

\title{Ensemble Performance Through the Lens of Linear Independence
of Classifier Votes in Data Streams}

\author{Enes~Bektas\,\orcidlink{0009-0007-5186-8634}~and~Fazli~Can\,\orcidlink{0000-0003-0016-4278}
\thanks{This work is under preparation for submission to the IEEE for possible publication. Copyright may be transferred without notice, after which this version may no longer be accessible.}
\thanks{E. Bektas and F. Can are with the Department of Computer Engineering, Bilkent University, Ankara, Turkey (e-mail: enes.bektas@bilkent.edu.tr; canf@cs.bilkent.edu.tr).}}



\maketitle

\begin{abstract}

Ensemble learning improves classification performance by combining multiple base classifiers. While increasing the number of classifiers generally enhances accuracy, excessively large ensembles can lead to computational inefficiency and diminishing returns. This paper investigates the relationship between ensemble size and performance through the lens of linear independence among classifier votes in data streams. We propose that ensembles composed of linearly independent classifiers maximize representational capacity, particularly under a geometric model. We then generalize the importance of linear independence to the weighted majority voting problem. By modeling the probability of achieving linear independence among classifier outputs, we derive a theoretical framework that explains the trade-off between ensemble size and accuracy. Our analysis leads to a theoretical estimate of the ensemble size  required to achieve a user-specified probability of linear independence. We validate our theory through experiments on both real-world and synthetic datasets using two ensemble methods, OzaBagging and GOOWE. Our results confirm that this theoretical estimate effectively identifies the point of performance saturation for robust ensembles like OzaBagging. Conversely, for complex weighting schemes like GOOWE, our framework reveals that high theoretical diversity can trigger algorithmic instability. Our implementation is publicly available to support reproducibility and future research\footnote{\href{https://github.com/EnesBektas/Ensemble-Performance-through-the-Lens-of-Linear-Independence-of-Classifier-Votes-in-Data-Streams-Arx}{Click here to visit GitHub repository.}}.
\end{abstract}


\section{INTRODUCTION}
\IEEEPARstart{E}{nsemble} learning is a widely used technique in machine learning and statistics for improving the performance and robustness of predictive models. An ensemble combines the predictions of multiple models, often referred to as "base classifiers" or "learners," to form a more accurate and resilient predictive model \cite{ASurveyonEnsembleLearningforDataStreamClassification,Asurveyonensemblelearning,Ensemblelearningfordatastreamanalysis:Asurvey,Ensemble-basedclassifiers,ErrorreductionthroughLearningMultipleDescriptions,ensemblemethodsinmachinelearning}. In recent years, with the exponential increase in data volume and complexity, ensemble models have gained popularity over single classifiers. The foundational principle
behind their success is the ability to mitigate the bias-variance
trade-off more effectively than individual models  \cite{Asurveyonensemblelearning}. This approach has been successfully applied in a variety of fields, including statistics, machine learning, pattern recognition, and knowledge discovery in databases \cite{ATaxonomyandShortReviewofEnsembleSelection}. Numerous studies have consistently demonstrated that ensembles generally outperform single classifiers in classification tasks \cite{Asurveyonensemblelearning, ErrorreductionthroughLearningMultipleDescriptions, ensemblemethodsinmachinelearning, EnsembleBasedSystemsInDecisionMaking, PopularEnsembleMethods:AnEmpiricalStudy}.

The efficacy of an ensemble is intrinsically linked to its construction methodology, which comprises two principal stages: the generation of diverse base classifiers and the subsequent combination of their predictions. Foundational generation methods documented in the literature include Bagging \cite{Bagging}, Boosting \cite{Boosting, Boosting2}, and Random Forests \cite{randomForests}. The methods for combining predictions are broadly classified into weighting schemes and meta-combination techniques \cite{Ensemble-basedclassifiers}. Weighting methods—such as Majority Voting, Performance Weighting, and Bayesian Combination—assign a specific weight to each base classifier, reflecting its contribution to the final prediction. In contrast, meta-combination techniques—such as Stacking and Arbiter Trees—employ a meta-learning approach, where a secondary model learns to optimally combine the base classifiers' outputs \cite{Asurveyonensemblelearning, Ensemble-basedclassifiers}. Among these, majority voting is a simple yet remarkably effective technique, particularly when the base classifiers are diverse and independent. The theoretical underpinnings that explain the effectiveness of majority voting are explored in detail in \cite{Applicationofmajorityvotingtopatternrecognition:ananalysisofitsbehaviorandperformance, Limitsonthemajorityvoteaccuracyinclassifierfusion, Theoreticalboundsofmajorityvotingperformanceforabinaryclassificationproblem, ATheoreticalAnalysisoftheLimitsofMajorityVotingErrorsforMultipleClassifierSystems}. This paper focuses specifically on weighting methods, with an emphasis on weighted majority voting.

While the fundamental concept of ensemble learning encourages the use of numerous classifiers, deploying an excessively large ensemble introduces significant computational overhead in terms of memory and processing time. Furthermore, the performance gains often diminish beyond a certain number of classifiers. Research into the relationship between ensemble size and performance has led to two main perspectives. The first posits that a smaller, carefully pruned subset of classifiers can perform comparably to the full ensemble, a concept often framed as "many could be similar to all" \cite{Anempiricalstudyofbuildingcompactensembles, Effectivepruningofneuralnetworkclassifierensembles, pruningadaptiveboosting, Pruninganddynamicschedulingofcost-sensitiveensembles}. The second perspective, characterized as "many could be better than all," argues that a subset of classifiers can even outperform the original, complete ensemble \cite{Ensemblingneuralnetworks:Manycouldbebetterthanall, Selectiveensembleofdecisiontrees}.

Conversely, other studies suggest that ensemble performance monotonically increases with ensemble size, although the marginal improvements progressively decrease \cite{Howlargeshouldensemblesofclassifiersbe?}. Our work aligns with this latter view, providing a theoretical foundation to explain the observed diminishing returns. To our knowledge, the theoretical principles governing this trade-off between ensemble size and predictive performance remain an open area of investigation. This paper introduces a theoretical framework to clarify this relationship, positing that the concept of linear independence among classifier votes is central to understanding and optimizing ensemble performance.

The contributions of this study are the following. We
\begin{itemize}
\item Theoretically demonstrate the importance of linear independence among classifier votes, providing a new algebraic perspective on ensemble diversity and its impact on performance.
\item Develop a probabilistic framework to model the trade-off between ensemble size and linear independence, culminating in two metrics to estimate the point of full diversity: a precise formula (\texttt{INC})  and a computationally efficient, closed-form approximation (\texttt{SINC}).
\item Validate our theoretical estimates through comprehensive experiments, showing they successfully identify the point of full performance saturation for robust ensembles (OzaBagging)  and, in contrast, reveal the instability and performance degradation of complex weighting schemes (GOOWE) at high diversity.
\end{itemize}

The remainder of this paper is organized as follows. Section II reviews prior work related to ensemble size, and introduces the geometric framework for our analysis. In Section III, we develop our core theory on ensemble construction. Section IV discusses the practical implications of this theory. Section V presents experimental results to validate our claims. Finally, Section VI provides concluding remarks.

\section{RELATED WORK}
This section reviews prior research relevant to our study. We first provide a broad overview of existing strategies for determining ensemble size, including ensemble selection, pruning, and other theoretical approaches. We then narrow our focus to the geometric framework for ensemble classification, discussing the conflicting recommendations that motivate our work.

\subsection{Determining Ensemble Size}
The optimal size of an ensemble has been a subject of extensive investigation, leading to various strategies for its determination. A significant body of research has focused on \textit{ensemble selection}, which involves constructing an optimal subset of classifiers from a larger pool. For instance, Ulaş et al. proposed an incremental construction method using accuracy, statistically significant improvement, and diversity as selection criteria \cite{Incrementalconstructionofclassifieranddiscriminantensembles}. Similarly, Xiao et al. introduced a dynamic selection approach that considers both accuracy and diversity, particularly for noisy data \cite{Adynamicclassifierensembleselectionapproachfornoisedata}, while Yang employed Q-statistics as a diversity measure alongside accuracy for classifier selection \cite{Classifiersselectionforensemblelearningbasedonaccuracyanddiversity}.

Another prominent research direction is \textit{ensemble pruning}, which aims to reduce the size of a pre-existing ensemble to improve computational efficiency without degrading predictive performance. The underlying principle is to first generate a large ensemble and subsequently eliminate redundant or underperforming members \cite{EnsemblePruningUsingSpectralCoefficients, pruningadaptiveboosting, Anefficientensemblepruningapproachbasedonsimplecoalitionalgames, Improveddiscreteartificialfishswarmalgorithmcombinedwithmargindistanceminimizationforensemblepruning, Optimizingthenumberoftreesinadecisionforesttodiscoverasubforestwithhighensembleaccuracyusingageneticalgorithm, Collective-agreement-basedpruningofensembles, dyned, OnTheFlyEnsemblePruningInEvolvingDataStreams}. However, a limitation of many pruning algorithms is their singular focus on performance, often neglecting the computational cost. Bhardwaj et al. highlighted this gap, arguing that the cost-effectiveness of an ensemble must be evaluated as a function of both its size and its accuracy \cite{Cost-effectivenessofclassificationensembles}.

Specific studies have also addressed the optimal number of trees in Random Forests. Latinne et al. applied the McNemar test to determine \textit{a priori} the minimum ensemble size required to achieve a performance level comparable to that of much larger forests \cite{LimitingtheNumberofTreesinRandomForests}. Oshiro et al. empirically demonstrated that while performance generally increases with the number of trees, the improvements become marginal beyond a certain threshold \cite{HowManyTreesinaRandomForest?}. Probst and Boulesteix provided a theoretical perspective, showing that the expected error rate of a random forest is not necessarily a monotonic function of the number of trees \cite{Totuneornottotunethenumberoftreesinrandomforest?}.

Other researchers have explored theoretical and analytical approaches to determine ensemble size. Bax analyzed the majority voting mechanism and concluded that any odd number of classifiers could be optimal, proposing a validation-based selection method \cite{Selectinganumberofvotersforavotingensemble}. Hernàndez-Lobato et al. developed a method to estimate the number of classifiers needed for a parallel ensemble's vote to approximate the vote of an infinite-sized ensemble with a user-defined confidence level \cite{Howlargeshouldensemblesofclassifiersbe?}. Fumera et al. conducted a theoretical analysis of bagging, modeling it as a linear combination of classifiers to derive the expected error as a function of ensemble size \cite{Atheoreticalandexperimentalanalysisoflinearcombinersformultipleclassifiersystems, ATheoreticalAnalysisofBaggingasaLinearCombinationofClassifiers}. In the context of data streams, Jackowski introduced a diversity measure based on classifier reactions to concept drift, which can also be used to control the ensemble size \cite{Newdiversitymeasurefordatastreamclassificationensembles}.

\begin{table}[h]
\caption{DEFINITION OF FREQUENTLY USED SYMBOLS}
\label{tab:freq}
\centering
\footnotesize
\begin{tabular}{p{2.4cm} p{5.5cm}}
\toprule
\textbf{Symbol} & \textbf{Definition} \\
\midrule
$\mathcal{D}$ = \{$I_{1}, I_{2}, ...$\} & The data stream. \\
$E$ = \{$C_{1},\dots, C_{n}$\} & The ensemble with $n$ component classifiers. \\
$I_k$ & An instance of data stream. \\
$m$ & Number of class labels in the data stream. \\
$n$ & Number of classifiers in the ensemble. \\
$o_k$ & The ideal vector in the form $(0, \dots, 0, 1, 0, \dots, 0)$. \\
$p_{l}$ & Probability of a classifier's vote being linearly dependent on an existing $l$-dimensional space. \\
$S_{i}$ & Vote vector of $i^{th}$ classifier in the ensemble. \\
$S_{ij}$ & $j^{th}$ component of $i^{th}$ classifier's vote. \\
$V_k$ & Vote of the ensemble for instance $I_k$ \\
$W_{i}$ & Weight of $i^{th}$ classifier in the ensemble. \\
\midrule
PLI & Probability of Linear Independence. \\
INC & Ideal Number of Classifiers (calculated from Theorem 2). \\
SINC & Simplified Ideal Number of Classifiers. \\
\bottomrule
\end{tabular}
\end{table}

\subsection{The Geometric Framework and the Ensemble Size Debate}

Our work builds upon a  geometric framework for data fusion, originally introduced by Wu and Crestani for information retrieval systems \cite{Ageometricframeworkfordatafusionininformationretrieval}. This framework was subsequently adapted for ensemble classification by Bonab and Can, who treated each classifier's vote as a vector in a high-dimensional space and used weighted majority voting \cite{lessismorecikm, lessIsMore}. Wu and Ding later extended this model to dataset-level classification \cite{ADataset-LevelGeometricFrameworkforEnsembleClassifiers}. We formally define the key components of this framework, with our notation summarized in Table \ref{tab:freq}, before discussing its relevance to our study.

In this geometric model, for a data stream \textit{$\mathcal{D}$} with \textit{m} class labels and an ensemble \textit{E} with \textit{n} classifiers, each classifier's vote on an instance $I_k$ is represented as a vector $S_i$ in an \textit{m}-dimensional space. Each vote vector is normalized such that the sum of its components is unity ($\sum_{j=1}^{m} S_{ij} = 1$). An ideal vector, $o_k$, is defined as a vector that indicates the true class label of instance $I_k$. The ensemble's final vote, $V$, is a weighted linear combination of the individual classifier votes. The optimal weights, $W_i$, are calculated to minimize a loss function defined as the Euclidean distance between the ideal vector $o_k$ and the ensemble's vote $V$ \cite{Ageometricframeworkfordatafusionininformationretrieval, lessIsMore}.

Within this framework, conflicting recommendations regarding the optimal ensemble size have emerged. Bonab and Can suggested that the number of classifiers should ideally equal the number of class labels (\textit{n = m}) for optimal weight assignment \cite{lessismorecikm, lessIsMore}. In contrast, Wu and Crestani proved that adding more classifiers generally improves performance, or at worst, leaves it unchanged \cite{Ageometricframeworkfordatafusionininformationretrieval}. This contradiction highlights a critical gap in understanding the principles that govern ensemble size. To resolve this ambiguity, we propose a probabilistic approach grounded in the concept of linear independence.

Table \ref{tab:related_summary} provides a summary of these prior theoretical frameworks and contrasts them with the approach proposed in this paper.

\begin{table*}[!t]
\caption{COMPARISON OF THEORETICAL FRAMEWORKS FOR ENSEMBLE SIZING}
\label{tab:related_summary}
\centering
\footnotesize
\begin{tabular}{l p{4cm} p{8cm}}
\toprule
\textbf{Approach} & \textbf{Core Concept / Model} & \textbf{Key Goal / Finding} \\
\midrule

Hernández-Lobato et al. \cite{Howlargeshouldensemblesofclassifiersbe?} & Statistical approximation & Estimates size $n$ needed to approximate the vote of an infinite-sized ensemble. \\
\addlinespace
Fumera et al. \cite{Atheoreticalandexperimentalanalysisoflinearcombinersformultipleclassifiersystems, ATheoreticalAnalysisofBaggingasaLinearCombinationofClassifiers} & Models bagging as a \newline linear combination & Derives the expected error as a function of ensemble size $n$. \\
\addlinespace
Bax \cite{Selectinganumberofvotersforavotingensemble} & Majority vote analysis & Proposes that any odd number can be optimal; suggests a validation-based selection. \\
\addlinespace
Wu \& Crestani \cite{Ageometricframeworkfordatafusionininformationretrieval} & Geometric framework & Proves that adding more classifiers generally improves performance (monotonicity); implies larger is better. \\
\addlinespace
Bonab \& Can \cite{lessismorecikm, lessIsMore} & Geometric framework & Minimizes Euclidean distance between ensemble vote and ideal vector; suggests a heuristic of $n=m$. \\
\addlinespace
\textbf{This Paper} & \textbf{Geometric framework \newline + Linear Algebra} & \textbf{Models the probability (PLI) of achieving $m$ linearly independent votes; provides a theoretical basis for performance saturation.} \\
\bottomrule
\end{tabular}
\end{table*}

\section{A THEORETICAL FRAMEWORK BASED ON LINEAR INDEPENDENCE}
Within the geometric framework, the process of calculating optimal weights reveals that the linear independence of classifier votes is a critical property. The ensemble's final prediction is a linear combination of its constituent classifiers' votes. Consequently, the degree of linear dependence among these vote vectors directly constrains the vector space spanned by the ensemble, which in turn determines its expressive power and predictive capacity.

Before detailing our theorems, it is important to clarify the relationship between our general theory and the data stream context. The following framework, based on the linear independence of vote vectors, is a general algebraic model. It is applicable to any weighted majority voting ensemble, whether the data is processed in a batch or as a stream.

However, we center our analysis on data streams for two critical reasons. First, the problem of ensemble size and computational overhead is acute in stream mining, where models must operate under strict resource constraints. Second, our instance-based analysis---which examines the properties of vote vectors for a given instance $I_k$---maps naturally to the online, instance-by-instance learning model required by data streams. Therefore, we develop this general theory to solve a problem to the data stream domain and validate it using standard stream classification methods and protocols. A summary of the theoretical contributions developed in this section is presented in Table \ref{tab:theorem_summary}.

\begin{table}[h] 
\caption{SUMMARY OF THEORETICAL CONTRIBUTIONS IN SECTION III}
\label{tab:theorem_summary}
\centering
\footnotesize 
\begin{tabular}{lp{6cm}} 
\toprule
\textbf{Theorem} & \textbf{Key Contribution} \\
\midrule
Theorem 1 & Establishes that $m$ linearly independent votes are sufficient for an ensemble to perfectly represent the true class label for a single instance. \\
\addlinespace 
Theorem 2 & Provides a formula to calculate the probability of achieving $m$ linearly independent votes within an ensemble of size $n$, given the dependence probabilities $p_l$. \\
\addlinespace
Theorem 3 & Proves that this probability converges to 1 as the ensemble size $n$ increases (assuming $p_l < 1$), theoretically justifying the diminishing returns in ensemble performance. \\
\bottomrule
\end{tabular}
\end{table}

The following theorem formalizes the importance of achieving a set of linearly independent votes.

\subsection{Perfect Classification Under Linear Independence}

\textbf{\textit{Theorem 1:}} For any instance $I_k$ from a data stream \textit{$\mathcal{D}$} with \textit{m} classes, if an ensemble $E$ contains at least \textit{m} classifiers that produce linearly independent vote vectors, then there exists a set of weights $W = \{W_{1}, ..., W_{n}\}$ such that $\sum_{i=1}^{n} W_i = 1$ and the resulting ensemble vote \textit{V} is identical to the ideal vector \textit{o}.

\begin{proof}
Let us first consider the case where the ensemble size \textit{n} is equal to the number of classes \textit{m}, and all \textit{m} classifiers provide linearly independent vote vectors $\{S_1, ..., S_m\}$ for instance $I_k$. Because these vectors form a basis for the \textit{m}-dimensional space, there exists a unique solution for the weights \textit{W} in the linear system:
\begin{gather}
    \begin{bmatrix}
        S_{11} & S_{21} & \cdots & S_{n1} \\
        S_{12} & S_{22} & \cdots & S_{n2} \\
        \vdots & \vdots & \ddots & \vdots \\
        S_{1m} & S_{2m} & \cdots & S_{nm}
    \end{bmatrix}
    \begin{bmatrix}
        W_{1} \\
        W_{2} \\
        \vdots \\
        W_{n}
    \end{bmatrix}
    =
    \begin{bmatrix}
        o_{1} \\
        o_{2} \\
        \vdots \\
        o_{m}
    \end{bmatrix}
    = o
\end{gather}
This system of equations is equivalent to $\sum_{i=1}^{n} W_i S_i = o$. Summing the components of the resulting vector equation from $j=1$ to $m$ gives:
\[ \sum_{j=1}^{m} \left( \sum_{i=1}^{n} W_i S_{ij} \right) = \sum_{j=1}^{m} o_j = 1 \]
By rearranging the order of summation, we get:
\[ \sum_{i=1}^{n} W_i \left( \sum_{j=1}^{m} S_{ij} \right) = 1 \]
Given that each vote vector is normalized such that $\sum_{j=1}^{m} S_{ij} = 1$ for all classifiers \textit{i}, the equation simplifies to:
\[ \sum_{i=1}^{n} W_i = 1 \]
For the case where \textit{$n > m$}, if a subset of \textit{m} classifiers provides linearly independent votes, a solution can be found by assigning the ideal weights to this subset as described above, while setting the weights of the remaining \textit{n - m} classifiers to zero. This completes the proof.
\end{proof}

\textbf{\textit{Discussion:}} Theorem 1 establishes that with \textit{m} linearly independent votes, an ensemble can perfectly classify any given instance $I_k$. However, the set of linearly independent classifiers may change from one instance to another. This dynamic nature necessitates a shift from a deterministic to a probabilistic perspective. We, therefore, seek to determine the probability of obtaining at least \textit{m} linearly independent votes within an ensemble of size \textit{n}.

\subsection{Probability of Achieving Linear Independence}

To model the ensemble construction process, we make a simplifying assumption that the probability of a classifier's vote being linearly dependent on the existing vote space depends solely on the current dimension of that space, and is uniform across all classifiers.

\textbf{\textit{Definition 1:}} Let $p_l$ denote the probability that a new classifier's vote vector lies within the subspace spanned by $l$ existing linearly independent vote vectors. In other words, $p_l$ is the probability that adding a new classifier fails to increase the dimension from $l$ to $l+1$. We assume $p_l$ is constant for any classifier added to an $l$-dimensional span. These probabilities can be estimated empirically from a given dataset.

\textbf{\textit{Definition 2:}} Let $\chi_k$ be the set of $(m-1)$-tuples of non-negative integers that sum to $k$:
\begin{multline*}
    \chi_k = \{ (x_1, \ldots, x_{m-1}) \mid \sum_{j=1}^{m-1} x_j = k; \ x_j \in \mathbb{N}_0 \}
\end{multline*}

\textbf{\textit{Theorem 2:}} Given the probabilities $p_{1}, ..., p_{m-1}$, the probability of obtaining at least \textit{m} linearly independent votes from an ensemble of \textit{n} classifiers ($n \ge m$) is given by:
\begin{equation}
P(n,m) = \left( \prod_{i=1}^{m-1} (1-p_{i}) \right) \left( \sum_{k=0}^{n-m} \sum_{(x_1, \ldots, x_{m-1}) \in \chi_k} \prod_{j=1}^{m-1} p_{j}^{x_j} \right)
\end{equation}

\begin{proof}
To understand the structure of this formula, consider the process of constructing a set of votes incrementally. We can model the growth of independence as a branching process, as illustrated in Figure \ref{fig:probtree}. The process of achieving \textit{m} linearly independent votes requires successfully increasing the dimension of the spanned subspace from 1 to \textit{m}. The probability of a new vote increasing the dimension from \textit{i} to \textit{i}+1 is $(1-p_i)$. Therefore, the probability of achieving \textit{m} linearly independent votes in the first \textit{m} attempts (i.e., with the first \textit{m} classifiers) is the product of these success probabilities, $\prod_{i=1}^{m-1}(1-p_{i})$. This term represents the base probability of the most direct path to success.

\begin{figure}[h]
    \centering
    \includegraphics[width=8cm]{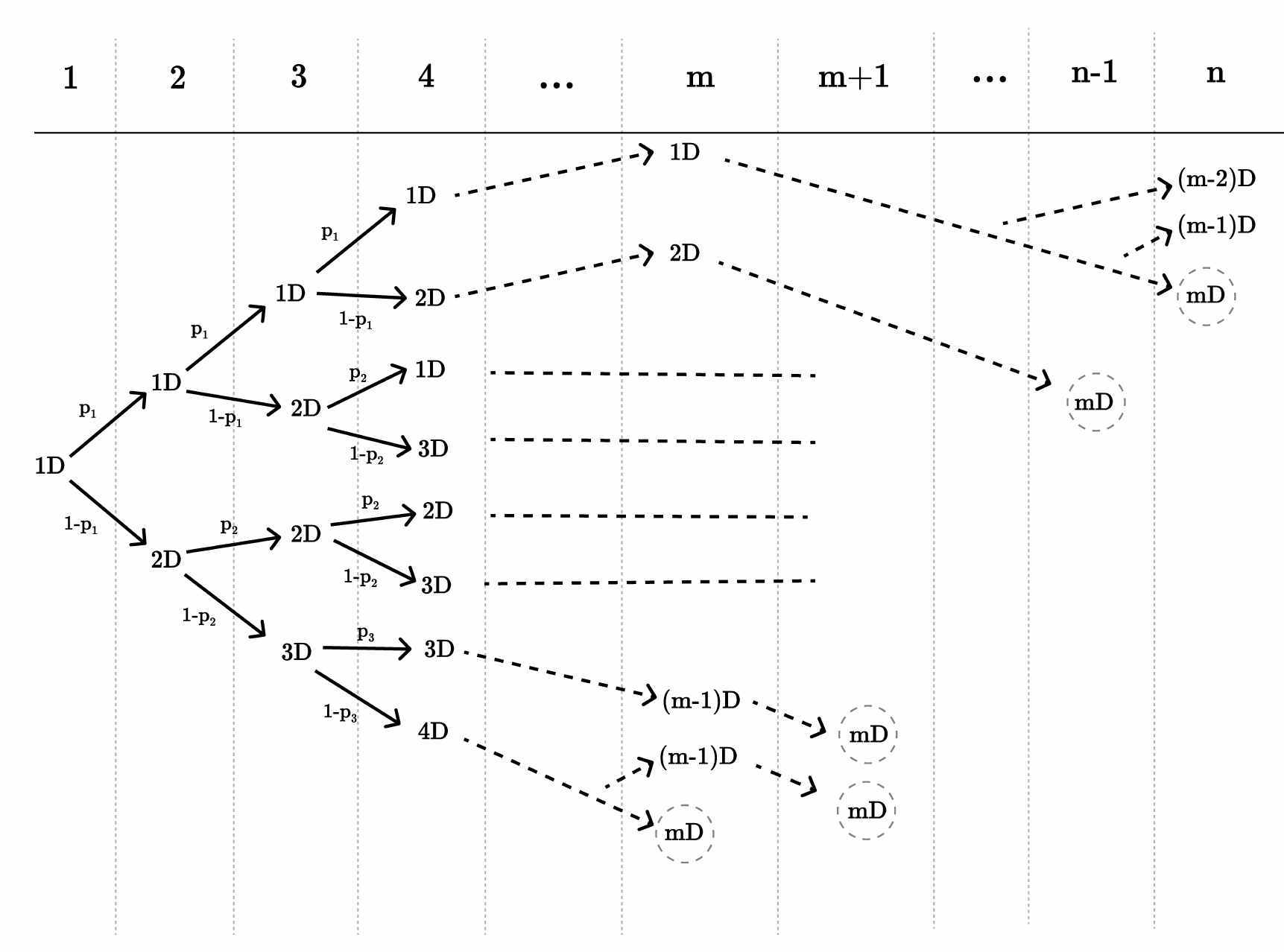}
    \caption{A conceptual probability tree illustrating the paths to achieving an \textit{m}-dimensional vote space with \textit{n} classifiers. Each branch represents the addition of a classifier, which either increases the dimension (D) or is linearly dependent.}
    \label{fig:probtree}
\end{figure}

If this is not achieved within the first \textit{m} classifiers, additional classifiers are needed. The term $\sum_{k=0}^{n-m} (\dots)$ accounts for all scenarios where success is achieved using up to $n$ classifiers. The outer sum over $k$ considers the total number of "wasted" classifiers—those that did not increase the dimension of the spanned space. The inner summation over $\chi_k$ considers all possible ways these $k$ dependent votes can be distributed. Specifically, $x_j$ represents the number of times a new vote was found to be dependent on the existing $j$-dimensional subspace. The term $\prod_{j=1}^{m-1} p_{j}^{x_j}$ is the probability of one such specific sequence of dependencies occurring. Summing over all possible combinations of these dependent votes gives the complete probability of achieving success within an ensemble of size \textit{n}.
\end{proof}

\textbf{\textit{Discussion:}} Theorem 2 provides a quantitative tool to evaluate the marginal benefit of adding a new classifier to the ensemble. It directly models the trade-off between ensemble size (\textit{n}) and the likelihood of achieving a sufficiently diverse set of votes. To understand the long-term behavior of this trade-off, we analyze the limit of this probability as the ensemble size grows infinitely large.

\subsection{Asymptotic Behavior of Ensemble Diversity}

\textbf{\textit{Theorem 3:}} Provided that $p_l \neq 1$ for all $l \in \{1, ..., m-1\}$, the probability of obtaining \textit{m} linearly independent votes converges to 1 as the ensemble size \textit{n} approaches infinity.
\begin{equation}
    \lim_{n \to \infty} P(n,m) = 1
\end{equation}

\begin{proof}
The convergence depends on the values of $p_l$.

\textit{Case 1: At least one $p_l = 1$.} If there exists an $l$ such that $p_l=1$, it is impossible to surpass an $l$-dimensional space. The term $\prod_{i=1}^{m-1}(1-p_{i})$ becomes 0, and thus the limit is 0.

\textit{Case 2: All $p_l = 0$.} If all $p_l=0$, every classifier is guaranteed to be linearly independent of the previous ones. The term $\prod_{i=1}^{m-1}(1-p_{i})$ becomes 1. Regarding the summation, any term where $k > 0$ implies at least one dependency occurred (some $x_j \ge 1$), causing the term to vanish since $p_j=0$. Therefore, the entire summation collapses to the single term for $k=0$ (where all $x_j=0$), which equals 1. This results in a total probability of 1 for any $n \ge m$.

\textit{Case 3: $0 \le p_l < 1$ for all $l$, with at least one $p_l > 0$.} We prove this by induction on \textit{m}. Let $S_m = \sum_{k=0}^{\infty} \sum_{(x_1, \ldots, x_{m-1}) \in \chi_k} \prod_{j=1}^{m-1} p_{j}^{x_j}$. We must show that $\left(\prod_{i=1}^{m-1}(1-p_{i})\right) S_m = 1$.

\textit{Base Case (m=2):} The expression becomes $(1-p_1) \sum_{k=0}^{\infty} p_1^k$. Since $0 \le p_1 < 1$, this is a geometric series which converges to $\frac{1}{1-p_1}$. Thus, $(1-p_1) \frac{1}{1-p_1} = 1$. The theorem holds.

\textit{Inductive Step:} Assume the theorem holds for $m-1=d$, i.e., $S_d = \frac{1}{\prod_{j=1}^{d-1}(1-p_j)}$. We must show it holds for $m=d+1$. Since the outer sum extends over all possible values of $k$ (from 0 to $\infty$), the constraint $\sum x_j = k$ covers the entire space of non-negative integers $\mathbb{N}_0^d$. This allows us to decouple $x_d$ from the other variables and factor out the terms involving $p_d$:
\begin{align*}
S_{d+1} &= \sum_{k=0}^{\infty} \sum_{(x_1, \ldots, x_{d}) \in \chi_{k}} \prod_{j=1}^{d} p_{j}^{x_j} \\ 
&= \sum_{x_d=0}^{\infty} p_d^{x_d} \left( \sum_{k'=0}^{\infty} \sum_{(x_1, \ldots, x_{d-1}) \in \chi_{k'}} \prod_{j=1}^{d-1} p_{j}^{x_j} \right) \\
&= \left( \sum_{x_d=0}^{\infty} p_d^{x_d} \right) \left( S_d \right) \\
&= \left(\frac{1}{1-p_d}\right) S_d
\end{align*}
By the inductive hypothesis, $S_d = \frac{1}{\prod_{j=1}^{d-1}(1-p_j)}$. Substituting this in, we get:
\[ S_{d+1} = \frac{1}{1-p_d} \cdot \frac{1}{\prod_{j=1}^{d-1}(1-p_j)} = \frac{1}{\prod_{j=1}^{d}(1-p_j)} \]
Therefore, $\left(\prod_{i=1}^{d}(1-p_{i})\right) S_{d+1} = 1$. The induction holds.
\end{proof}

\textbf{\textit{Discussion:}} Theorems 2 and 3 provide theoretical justification for the empirical observation that larger ensembles tend to perform better, but with diminishing returns. Theorem 3 proves that, under the reasonable assumption that no feature space is inherently impassable ($p_l \neq 1$), achieving a full-rank set of votes is a probabilistic certainty given a large enough ensemble. This convergence allows us to reframe the problem of finding the optimal ensemble size: instead of seeking a universal number, we can determine the minimum ensemble size \textit{n} required to achieve a desired target probability (e.g., 99.9\%) of obtaining \textit{m} linearly independent votes. This provides a principled and practical method for guiding ensemble construction.

\section{PRACTICAL IMPLICATIONS}
The preceding theoretical development has direct practical applications for ensemble design. A foundational principle in ensemble learning is that diversity among base classifiers is essential for robust performance \cite{Ensemblediversitymeasuresandtheirapplicationtothinning}. Our framework contributes to this understanding by providing a formal, algebraic interpretation and a mathematical measure for diversity: The linear independence of classifier vote vectors. We claim that a higher probability of achieving linear independence directly corresponds to increased ensemble diversity and, consequently, to a more powerful ensemble, up to the point of performance saturation.

Theorem 2 offers a quantitative metric to assess this form of diversity. This allows for a quantitative approach to determining the required ensemble size, \textit{n}, moving beyond the fixed heuristic of $n=m$ previously derived from this geometric framework \cite{lessismorecikm, lessIsMore}. Our methodology involves specifying a desired confidence threshold, \textit{T}, which represents the minimum acceptable probability of obtaining \textit{m} linearly independent votes. By applying the formula from Theorem 2, one can then calculate the minimum number of classifiers, \textit{n}, required to ensure this probability meets or exceeds the threshold \textit{T}. This procedure enables the construction of ensembles that are provably diverse with a user-specified degree of confidence, providing a solution to the trade-off between ensemble size and performance.

The remainder of this section details these practical implications. First, Subsection A establishes the generalizability of our theorems beyond the specific geometric framework, demonstrating their relevance to all weighted majority voting schemes. Subsection B then provides a worked example illustrating how to determine the minimum ensemble size for a given probability threshold. Following this, Subsection C analyzes the computational complexity of the formula presented in Theorem 2. To address the potential for high computational costs, Subsection D introduces a simplified, closed-form approximation of the formula. Finally, Subsection E presents a practical algorithm for empirically estimating the crucial $p_l$ parameters from a dataset, making the entire framework applicable to real-world problems.

\subsection{Generalizing Beyond the Geometric Model}
While our theorems were derived within a specific geometric framework, their central conclusion—the critical importance of linear independence—extends to the broader class of all weighted majority voting ensembles.

In any weighted majority voting scheme, the ensemble's final decision is based on an aggregated vote vector, $V$, which is a linear combination of the vote vectors, $\{S_1, S_2, ..., S_n\}$, produced by the $n$ base classifiers. Each classifier's output, whether a probability distribution across classes or a one-hot encoding of its predicted class, can be naturally represented as a vector within an $m$-dimensional class space. The set of all possible aggregated vote vectors, $V$, that the ensemble can produce is therefore confined to the span of the base classifiers' vote vectors: $V \in \operatorname{span}(\{S_1, S_2, ..., S_n\})$.

The classification rule in such a scheme is to select the class corresponding to the index of the maximum component in the vector $V$. The fundamental goal is to select weights such that this maximum component aligns with the true class label of a given instance. However, the ability of the ensemble to achieve this is fundamentally constrained by the dimensionality of its spanned space.

If the set of base classifiers fails to produce at least $m$ linearly independent vote vectors, their span will be a subspace of dimension less than $m$. Consequently, there will exist regions of the $m$-dimensional class space that are unreachable by the ensemble. For an instance whose true class corresponds to an outcome in one of these unreachable regions, it is impossible to find a set of weights that will result in a vector $V$ where the true class's component is maximal. Regardless of the weighting strategy, the ensemble is representationally incapable of making the correct prediction, leading to an irreducible error. Common constraints on the weights, such as requiring them to sum to one ($\sum W_i = 1$), do not alter this conclusion, as it is just a scaling factor that does not change the index of the maximum value.

This demonstrates that achieving $m$ linearly independent votes is a necessary condition for an ensemble to have the capacity to correctly classify all possible outcomes. As established by Theorem 1, the presence of $m$ linearly independent votes guarantees this representational capacity is met. Therefore, the principles of linear independence and the probabilistic framework developed in this paper are not confined to the initial geometric model but are of universal importance to the weighted majority voting problem.

\subsection{Example: Computing Minimum Ensemble Size for Target Confidence}
To demonstrate the practical utility of our theoretical framework, we present a concrete example of determining the minimum ensemble size required to meet a specific confidence level.

Consider a binary classification problem where the number of classes, $m$, is 2. Let us assume that through empirical analysis of a given dataset, the probability of a new classifier's vote being linearly dependent on a single existing vote vector is found to be $p_1 = 0.5$. Our objective is to calculate the minimum ensemble size, $n$, required to achieve a probability of at least 99\% ($T=0.99$) of obtaining two linearly independent votes.

For the case where $m=2$, the general probability formula from Theorem 2 simplifies to the sum of a geometric series:
\[ P(n, 2) = (1 - p_1) \sum_{k=0}^{n-2} p_1^k \]

We apply this formula iteratively, increasing the ensemble size $n$ and calculating the resulting probability until it meets or exceeds our target threshold of 0.99. With $p_1 = 0.5$:
\begin{itemize}
    \item For $n=2$: $P(2,2) = (1-0.5) \cdot (0.5^0) = 0.5$
    \item For $n=3$: $P(3,3) = (1-0.5) \cdot (0.5^0 + 0.5^1) = 0.75$
    \item For $n=4$: $P(4,4) = (1-0.5) \cdot (0.5^0 + 0.5^1 + 0.5^2) = 0.875$
\end{itemize}
Continuing this process, we find that for an ensemble of size $n=7$, the probability is approximately $0.9844$, which is still below our target. Incrementing the size one final time:
\begin{itemize}
    \item For $n=8$: $P(8,8) = (1-0.5) \cdot \sum_{k=0}^{6} 0.5^k \approx 0.9922$
\end{itemize}
Since $P(8,2) > 0.99$, we conclude that a minimum of 8 classifiers are required to satisfy the condition of having two linearly independent votes with at least 99\% confidence. This example illustrates how our framework provides a methodology for ensemble sizing.

\subsection{Computational Complexity Analysis}
This section analyzes the computational complexity of iteratively calculating the probability derived in Theorem 2. The primary computational burden lies within the nested summation:
\begin{equation}
\label{eq:summation}
\sum_{k=0}^{n-m} \left( \sum_{(x_1, \ldots, x_{m-1}) \in \chi_k} \prod_{j=1}^{m-1} p_{j}^{x_j} \right)
\end{equation}

To determine the probability for an ensemble of size $n$, we must evaluate the inner sum for each value of $k$ from $0$ to $n-m$. The complexity of this operation is determined by the number of terms in the inner sum, which corresponds to the number of ways the integer $k$ can be expressed as the sum of $m-1$ non-negative integers. This is a classic combinatorial problem known as "stars and bars." The number of terms for a given $k$ is given by the multiset coefficient:
\begin{equation}
|\chi_k| = \binom{k + (m-1) - 1}{(m-1) - 1} = \binom{k+m-2}{m-2}
\end{equation}

The total number of product terms that must be computed to find the probability for an ensemble of size $n$ is the sum of these binomial coefficients over the full range of $k$. By applying the hockey-stick identity, this summation simplifies to a single binomial coefficient:
\begin{equation}
\text{Total Terms} = \sum_{k=0}^{n-m} \binom{k+m-2}{m-2} = \binom{n-1}{m-1}
\end{equation}

The complexity is therefore determined by the asymptotic behavior of this resulting term. The binomial coefficient $\binom{n-1}{m-1}$ can be expanded as:
\[ \binom{n-1}{m-1} = \frac{(n-1)(n-2)\cdots(n-m+1)}{(m-1)!} \]
For a fixed number of classes $m$, the denominator $(m-1)!$ is a constant, while the numerator is a polynomial in $n$ of degree $m-1$. This leads to an overall time complexity of $O(n^{m-1})$ for calculating the probability up to an ensemble size of $n$. 

As this analysis indicates, the direct application of the formula becomes computationally intensive for problems with a large number of classes ($m$), necessitating the development of a more efficient computational approach.

\subsection{Closed-Form Approximation for Uniform Dependence Probability}
The $O(n^{m-1})$ complexity of the formula derived from Theorem 2 renders it computationally expensive for datasets with a large number of classes ($m$). To overcome this barrier, we introduce a simplifying assumption: the probability of linear dependency is uniform across all dimensions. That is, we assume $p_l = p$ for all $l \in \{1, ..., m-1\}$, where $p$ represents a single, average probability that any new classifier's vote will be linearly dependent on the existing vote space.

This assumption allows us to transform the original nested summation into a closed-form expression. The derivation proceeds as follows. First, we substitute the uniform probability $p$ into the general formula:
\begin{align*}
P(n,m) &= (1-p)^{m-1} \sum_{k=0}^{n-m} \left( \sum_{(x_1, \ldots, x_{m-1}) \in \chi_k} \prod_{j=1}^{m-1} p^{x_j} \right) \\
&= (1-p)^{m-1} \sum_{k=0}^{n-m} p^k \left( \sum_{(x_1, \ldots, x_{m-1}) \in \chi_k} 1 \right)
\end{align*}
The inner sum is now simply the number of terms, $|\chi_k|$, which is given by the binomial coefficient $\binom{k+m-2}{m-2}$. This yields:
\begin{equation}
P(n,m) = (1-p)^{m-1} \sum_{k=0}^{n-m} \binom{k+m-2}{m-2} p^k
\end{equation}
The key insight for further simplification is to recognize the term $\binom{k+m-2}{m-2} p^k$ as being related to the derivative of a geometric series. Specifically, the polynomial part can be generated by differentiation:
\[
\begin{split}
\binom{k+m-2}{m-2} p^k &= \frac{(k+m-2)!}{k!(m-2)!}p^k \\
&= \frac{1}{(m-2)!} \frac{d^{m-2}}{dp^{m-2}} (p^{k+m-2})
\end{split}
\]
By substituting this back into the summation and interchanging the sum and derivative operators, we get:
\begin{align*}
P(n,m) &= \frac{(1-p)^{m-1}}{(m-2)!} \sum_{k=0}^{n-m} \frac{d^{m-2}}{dp^{m-2}}(p^{k+m-2}) \\
&= \frac{(1-p)^{m-1}}{(m-2)!} \frac{d^{m-2}}{dp^{m-2}} \left( \sum_{k=0}^{n-m} p^{k+m-2} \right)
\end{align*}
The summation inside the derivative is now a standard geometric series. Summing this series gives the final closed-form expression:
\begin{equation}
P(n,m) = \frac{(1-p)^{m-1}}{(m-2)!} \frac{d^{m-2}}{dp^{m-2}} \left( p^{m-2} \frac{1-p^{n-m+1}}{1-p} \right)
\end{equation}
This simplified formula allows the probability to be calculated without explicit summation over $n$. The complexity is now dominated by the $(m-2)$-th derivative of a rational function, which, for a fixed $m$, is independent of the ensemble size $n$. 

This reduces the complexity from $O(n^{m-1})$ to $O(1)$ with respect to $n$, making the calculation efficient even for very large ensembles.

\subsection{Empirical Estimation of Linear Dependence Probabilities ($p_l$)}

\begin{algorithm}
\caption{Algorithm for Empirically Estimating $p_l$ Values\protect}
\label{alg:p_calculation}
\begin{algorithmic}[1]
\State Let $m \gets$ number of class labels
\State $counts\_dependent \gets \text{zero array of size } m-1$
\State $counts\_total \gets \text{zero array of size } m-1$
\For{each instance $I$ in the dataset}
    \State $voteMatrix \gets n \times m$ matrix of classifier votes for $I$
    \State $curMatrix \gets \text{empty matrix}$
    \State $prevRank \gets 0$
    \For{$i = 1 \to n$}
        \State Append $i$-th row of $voteMatrix$ to $curMatrix$
        \State $curRank \gets \text{rank}(curMatrix)$
        \If{$curRank < m$}
            \State $counts\_total[prevRank] \gets counts\_total[prevRank] + 1$ \label{line:total_inc}
            \If{$curRank == prevRank$} \label{line:check_rank_equal}
                \State $counts\_dependent[prevRank] \gets counts\_dependent[prevRank] + 1$ \label{line:dependent_inc}
            \EndIf
        \Else
            \State \textbf{break} \Comment{Matrix has reached full rank} \label{line:break_full_rank}
        \EndIf
        \State $prevRank \gets curRank$ \label{line:update_prev_rank}
    \EndFor
\EndFor
\State $p\_values \gets \text{zero array of size } m-1$ \label{line:init_p_values}
\For{$l = 0 \to m-2$} \label{line:loop_calc_p}
    \If{$counts\_total[l] > 0$} \label{line:check_total_zero}
        \State $p\_values[l] \gets counts\_dependent[l] / counts\_total[l]$ \label{line:calc_p_ratio}
    \Else
        \State $p\_values[l] \gets 1.0$ \Comment{If dimension $l+1$ was never reached} \label{line:set_p_one}
    \EndIf
\EndFor
\State \Return $p\_values$ \Comment{$p\_values[l]$ is the estimate for $p_{l+1}$} \label{line:return_p}
\end{algorithmic}
\end{algorithm}

To apply the theoretical framework to real-world data, the linear dependence probabilities, $p_l$, must be estimated. We propose an empirical method, detailed in Algorithm \ref{alg:p_calculation}, to calculate these values from a given dataset and ensemble. The fundamental principle is to simulate the incremental construction of a basis of vote vectors for each data instance and to aggregate the observed frequencies of linear dependence at each dimensional step.

The algorithm initializes two arrays, \texttt{counts\_dependent} and \texttt{counts\_total} (Lines 2-3), to store counts for each potential dimension from 0 to $m-2$. It then iterates through every instance $I$ in the dataset (Line 4). For each instance, it gathers the $n \times m$ matrix of vote vectors (Line 5) and initializes an empty matrix \texttt{curMatrix} and a rank tracker \texttt{prevRank} (Lines 6-7). The core logic resides in the inner loop (Lines 8-20), which progressively builds \texttt{curMatrix} by adding one vote vector at a time (Line 9) and computing its rank, \texttt{curRank} (Line 10).

Crucially, each time a vector is added when the previous rank was \texttt{prevRank}, the algorithm increments the total count for attempts to expand that dimension (\texttt{counts\_total[prevRank]}, Line \ref{line:total_inc}). If adding the vector does not increase the rank (\texttt{curRank == prevRank}, Line \ref{line:check_rank_equal}), it signifies linear dependence, and the corresponding \texttt{counts\_dependent[prevRank]} is incremented (Line \ref{line:dependent_inc}). The loop breaks early if the matrix reaches full rank $m$ (Line \ref{line:break_full_rank}). The \texttt{prevRank} is updated at the end of each iteration (Line \ref{line:update_prev_rank}).

After processing all instances, the algorithm calculates the final probabilities (Lines \ref{line:init_p_values}-\ref{line:return_p}). For each dimension $l$ from 0 to $m-2$, the probability $p_{l+1}$ (stored in \texttt{p\_values[l]}) is computed as the ratio of dependent counts to total counts (Line \ref{line:calc_p_ratio}), provided that attempts were actually made to expand the $(l+1)$-dimensional space (Line \ref{line:check_total_zero}). If a dimension $(l+1)$ was never reached (i.e., \texttt{counts\_total[l]} is 0), $p_{l+1}$ is conservatively set to 1.0 (Line \ref{line:set_p_one}). The resulting array \texttt{p\_values} contains the empirical estimates for $p_1, \ldots, p_{m-1}$. This procedure provides the necessary empirical parameters for applying our theoretical model.

\section{EXPERIMENTS}
In this section, we present a series of experiments designed to empirically validate our theoretical framework. The primary objectives are threefold: 1) to investigate the relationship between the Probability of Linear Independence (PLI) and ensemble accuracy, 2) to demonstrate the practical utility of our proposed method for determining ensemble size, and 3) to explore the behavior of our framework across different ensemble methods and dataset characteristics.

To this end, Subsection A details our experimental setup, including the ensemble methods, datasets, and evaluation protocols used. Subsection B presents the empirical results, and Subsection C provides a detailed discussion of these findings, validating our theoretical model as a practical heuristic for ensemble sizing.

\subsection{Experimental Setup}
\subsubsection{\textit{Ensemble Methods and Base Learners}}
Two distinct ensemble methods were employed: OzaBagging \cite{ozabagging} and GOOWE \cite{goowe}. OzaBagging is a variant of bootstrap aggregating that combines classifier votes via majority voting. It is important to note that majority voting is a special case of weighted majority voting where all weights are equal, thus making OzaBagging a suitable method for validating our theoretical claims regarding the generalizability of linear independence across weighted voting schemes. GOOWE was selected as it directly implements the geometric weighting framework that inspired our theoretical development. To support the theoretical assumption of consistent probabilities of dependence ($p_l$) across classifiers, all ensembles are constructed using Hoeffding Trees as the base classifier \cite{hoeffding}.

\subsubsection{\textit{Datasets and Evaluation}}
Performance was assessed on twelve datasets, detailed in Table \ref{tab:specTable}. These datasets were selected to provide a diverse and representative evaluation, covering a variety of data types and patterns. This includes six real-world benchmarks chosen from different contexts (e.g., Airlines, Poker, Electricity) to ensure variety in the number of attributes, instances, and class label distributions. This real-world set is complemented by six synthetic datasets, generated using the scikit-multiflow library's random RBF generator \cite{skmultiflow}. This synthetic data allows us to systematically control for and isolate the effect of the number of classes ($m$), a key parameter in our theoretical framework.

\begin{table}[ht]
\captionsetup{aboveskip=5pt}
\centering
\footnotesize
\caption[specTable]{DATASET SPECIFICATIONS. THE UPPER HALF IS REAL-LIFE, THE LOWER HALF IS SYNTHETIC DATASETS}
\label{tab:specTable}
\begin{tabular}{l r r r}
  \toprule
  \textbf{Dataset} & \textbf{\#Instance} & \textbf{\#Attr.} & \textbf{\#Class Labels (m)} \\
  \midrule
  Airlines & 539,383 & 7 & 2 \\
  Click Prediction & 399,482 & 11 & 2 \\
    Electricity & 45,312 & 6 & 2 \\
  Covtype & 581,012 & 54 & 7 \\
  Poker & 829,201 & 10 & 10 \\
  Rialto & 82,250 & 27 & 10 \\
  \midrule
    RBF2 & 1,000,000 & 20 & 2 \\
  RBF4 & 1,000,000 & 20 & 4 \\
  RBF8 & 1,000,000 & 20 & 8 \\
  RBF16 & 1,000,000 & 20 & 16 \\
  RBF32 & 1,000,000 & 20 & 32 \\
  RBF64 & 1,000,000 & 20 & 64 \\
  \bottomrule
\end{tabular}
\end{table}

\begin{table*}[!t]
\centering
\caption{VALIDATION OF THEORETICAL ENSEMBLE SIZE AS A PRACTICAL HEURISTIC}
\label{tab:inc_sinc_combined_vertical}
\footnotesize
\begin{tabular*}{\textwidth}{@{\extracolsep{\fill}} llcccccc}
\toprule
\textbf{Dataset} & \textbf{Method} & \textbf{m} & \textbf{SINC} & \textbf{INC} & \textbf{$n_{INC}$} & \textbf{Accuracy @$n_{INC}$ (\% of Max)} & \textbf{Correlation} \\
\midrule
    Airlines & OzaBagging & 2 & 3 & 3 & 4 & 98.4969 & 0.775 \\
Airlines & GOOWE & 2 & 4 & 4 & 4 & 94.8549 & 0.605 \\
\addlinespace
Click Prediction & OzaBagging & 2 & 3 & 3 & 4 & 99.9181 & 0.994 \\
Click Prediction & GOOWE & 2 & 3 & 3 & 4 & 99.9545 & 0.824 \\
\addlinespace
Electricity & OzaBagging & 2 & 4 & 4 & 4 & 99.0396 & 0.797 \\
Electricity & GOOWE & 2 & 4 & 4 & 4 & 99.3130 & 0.337 \\
\addlinespace
Covtype & OzaBagging & 7 & 25 & 33 & 32 & 99.7248 & 0.940 \\
Covtype & GOOWE & 7 & 33 & 42 & 32 & 99.7320 & 0.971 \\
\addlinespace
Poker & OzaBagging & 10 & 30 & 39 & 32 & 99.6704 & 0.698 \\
Poker & GOOWE & 10 & 50 & 61 & 64 & 97.2533 & 0.034 \\
\addlinespace
Rialto & OzaBagging & 10 & 22 & 22 & 16 & 99.0780 & 0.418 \\
Rialto & GOOWE & 10 & 24 & 26 & 32 & 95.1252 & 0.897 \\
\midrule 
RBF2 & OzaBagging & 2 & 8 & 8 & 8 & 99.5581 & 0.913 \\
RBF2 & GOOWE & 2 & 5 & 5 & 4 & 87.9696 & 0.714 \\
\addlinespace
RBF4 & OzaBagging & 4 & 7 & 8 & 8 & 99.4517 & 0.905 \\
RBF4 & GOOWE & 4 & 7 & 7 & 8 & 89.0144 & 0.687 \\
\addlinespace
RBF8 & OzaBagging & 8 & 13 & 14 & 16 & 99.2480 & 0.885 \\
RBF8 & GOOWE & 8 & 13 & 13 & 16 & 97.1040 & 0.857 \\
\addlinespace
RBF16 & OzaBagging & 16 & 29 & 34 & 32 & 100.0000 & 0.709 \\
RBF16 & GOOWE & 16 & 29 & 32 & 32 & 99.4795 & 0.912 \\
\addlinespace
RBF32 & OzaBagging & 32 & 56 & 66 & 64 & 100.0000 & 0.307 \\
RBF32 & GOOWE & 32 & 55 & 65 & 64 & 100.0000 & 0.448 \\
\addlinespace
RBF64 & OzaBagging & 64 & -- & -- & -- & -- & -- \\
RBF64 & GOOWE & 64 & -- & -- & -- & -- & -- \\
\bottomrule
\end{tabular*}
\parbox{\textwidth}{\vspace{5pt}\footnotesize \textit{Note:} INC (Ideal Number of Classifiers) and SINC (Simplified Ideal Number of classifiers) are the theoretical ensemble sizes calculated to achieve a PLI threshold of 0.9999. $n_{INC}$ is the closest tested ensemble size (from $n=2$ to $n=128$) to the \texttt{INC} value. The Accuracy @$n_{INC}$ (\% of Max) column shows the average accuracy achieved at $n_{INC}$, expressed as a percentage of the maximum average accuracy observed for that dataset. Correlation is the Pearson correlation between PLI and Average Accuracy across all tested ensemble sizes.}
\end{table*}

\begin{figure*}[p]
    \centering
    \includegraphics[width=\textwidth]{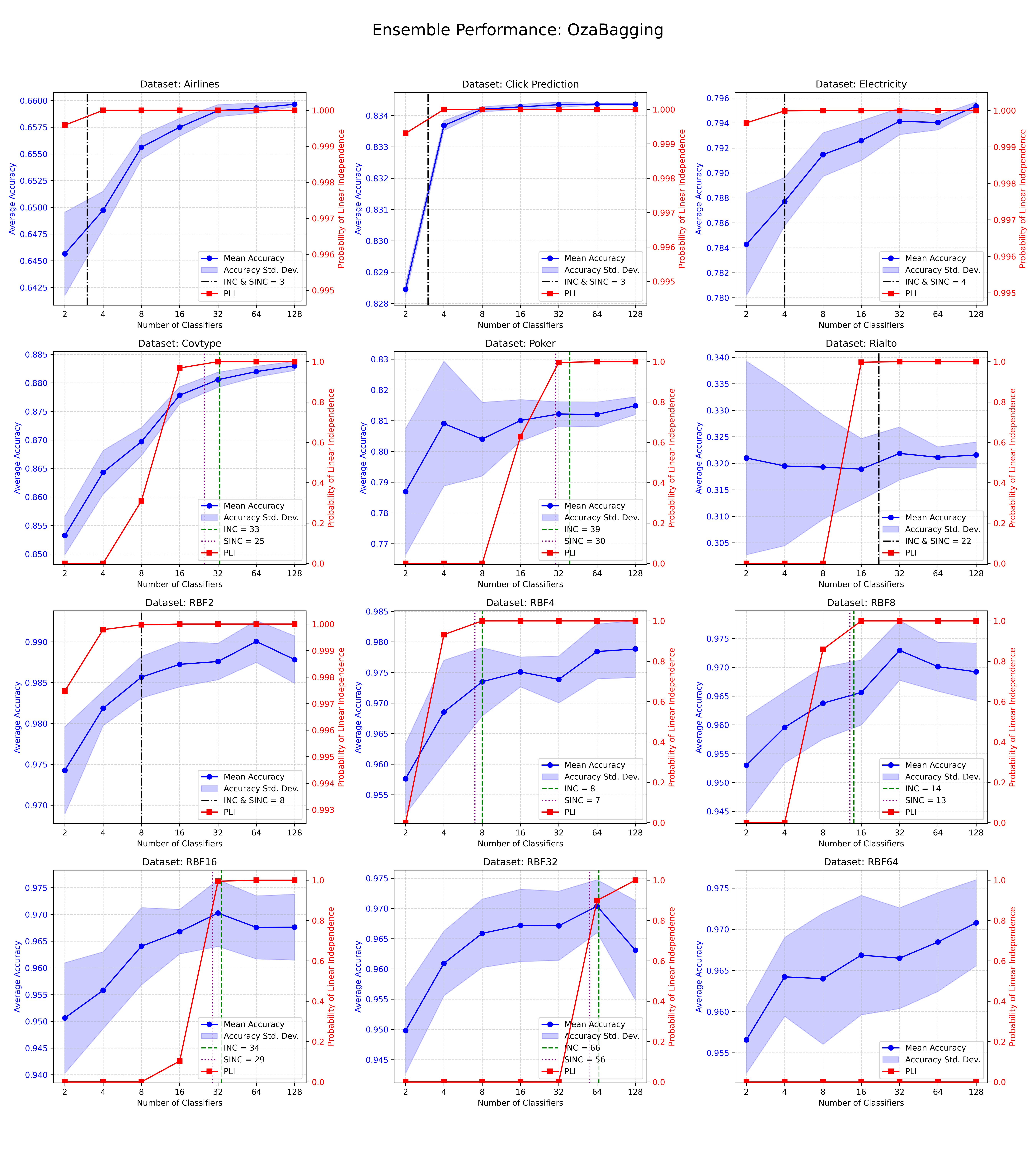}
\caption{Performance of OzaBagging. Each subplot shows the average accuracy (blue line, left y-axis) with standard deviation represented by the shaded area, and the Probability of Linear Independence (PLI) (red, right y-axis) as a function of ensemble size. \textbf{Note that PLI is 0 for all $n < m$ (where $m$ is the number of classes), as $m$ linearly independent vectors cannot be obtained from an ensemble smaller than $m$.} The vertical dashed lines indicate the theoretically derived Ideal Number of Classifiers (INC, green; SINC, purple) for a PLI threshold of 0.9999. Note: If INC and SINC coincide, only a single black line is shown. No INC/SINC lines are shown for RBF64 as the PLI remains 0 for all tested ensemble sizes.}
    \label{fig:ozabag}
\end{figure*}

\begin{figure*}[p]
    \centering
    \includegraphics[width=\textwidth]{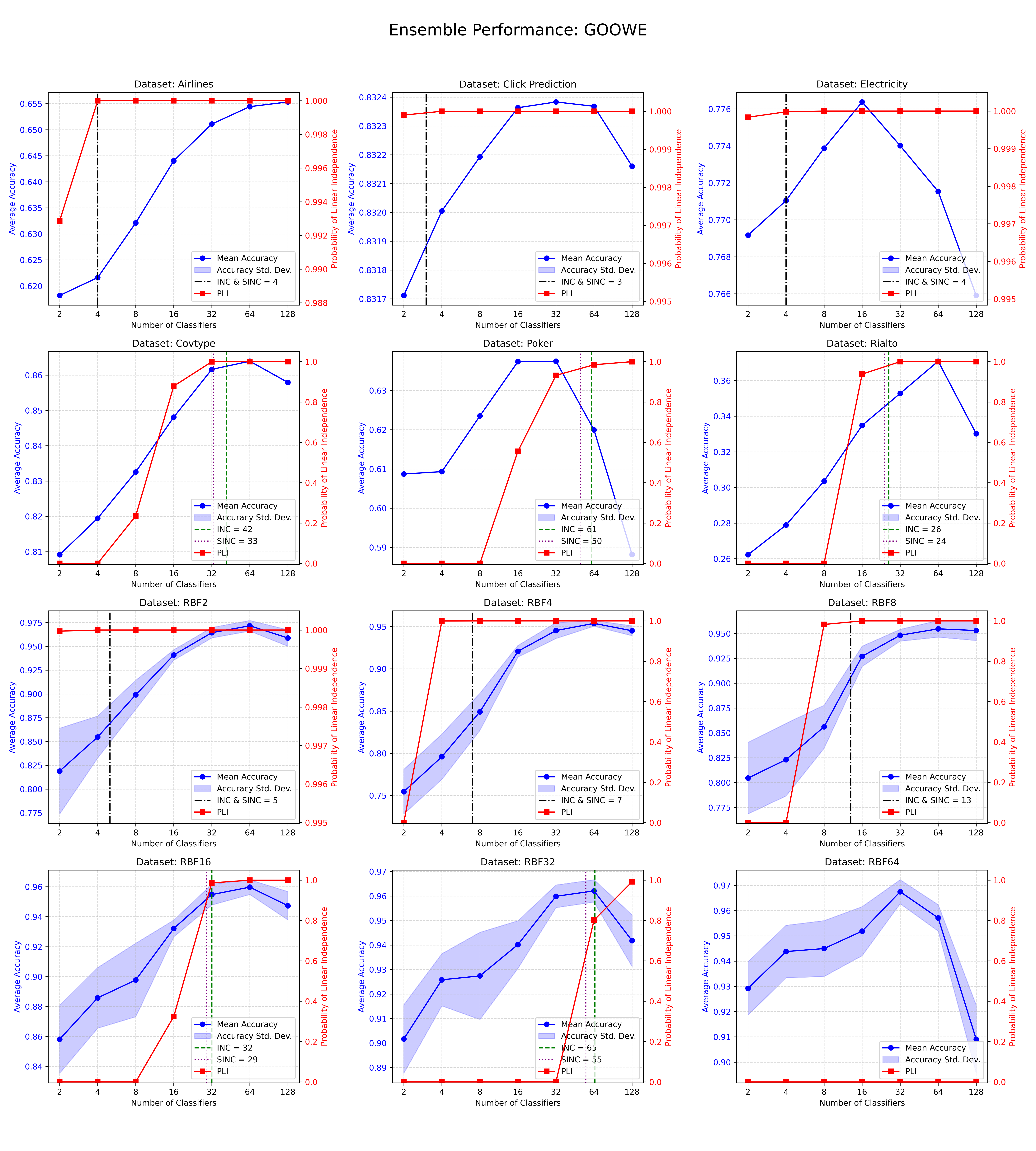}
    \caption{Performance of GOOWE. Each subplot shows the average accuracy (blue line, left y-axis) with standard deviation represented by the shaded area, and the Probability of Linear Independence (PLI) (red, right y-axis) as a function of ensemble size. \textbf{Note that PLI is 0 for all $n < m$ (where $m$ is the number of classes), as $m$ linearly independent vectors cannot be obtained from an ensemble smaller than $m$.} The vertical dashed lines indicate the theoretically derived Ideal Number of Classifiers (INC, green; SINC, purple) for a PLI threshold of 0.9999. Note: If INC and SINC coincide, only a single black line is shown. No INC/SINC lines are shown for RBF64 as the PLI remains 0 for all tested ensemble sizes. Note the instances of performance degradation at larger ensemble sizes, which contrasts with the behavior of OzaBagging.}
    \label{fig:goowe}
\end{figure*}

All experiments were evaluated using the prequential (interleaved test-then-train) methodology, which is standard for data stream classification. For each dataset and ensemble method, the experiments were repeated 10 times with different random seeds, and we report the average accuracy and standard deviation.

\subsubsection{\textit{PLI and Ideal Ensemble Size Calculation}}
The probabilities of linear dependence, $p_l$, were empirically estimated for each dataset and ensemble size using Algorithm 1. The Probability of Linear Independence (PLI) plotted for an ensemble of size $n$ was calculated using the specific $p_l$ values measured at that size. We also calculated two theoretical estimates for the "Ideal Number of Classifiers" required to achieve a PLI of 0.9999. We selected this threshold ($T=0.9999$) to identify the point of practical convergence, ensuring that the ensemble has effectively maximized its algebraic representational capacity.

\begin{itemize}
    \item INC (Ideal Number of Classifiers): Calculated using Theorem 2, with $p_l$ values that were averaged across all tested ensemble sizes (from 2 to 128) to provide a single, robust estimate for the dataset.
    \item SINC (Simplified INC): Calculated using our simplified formula, with a single probability $p$ derived from the average of all $p_l$ values.
\end{itemize}
All experimental code is publicly available for reproducibility\footnote{\href{https://github.com/EnesBektas/Ensemble-Performance-through-the-Lens-of-Linear-Independence-of-Classifier-Votes-in-Data-Streams-Arx}{Click here to visit GitHub repository.}}.

\subsection{Results}
The experimental results for OzaBagging and GOOWE are presented visually in Figure \ref{fig:ozabag} and Figure \ref{fig:goowe}, respectively. We begin with OzaBagging to establish a baseline under equal-weighted majority voting, followed by GOOWE to examine performance under a complex, geometry-driven weighting scheme.

Visually, OzaBagging (Figure \ref{fig:ozabag}) displays a monotonic increase in accuracy as the ensemble size ($n$) grows, typically stabilizing at a high value. In contrast, GOOWE (Figure \ref{fig:goowe}) exhibits more complex behavior; while accuracy initially rises, it frequently plateaus earlier or degrades at larger ensemble sizes (e.g., Poker, RBF32). In both figures, the Probability of Linear Independence (PLI), shown in red, consistently rises from 0 toward 1 as $n$ increases, except for the high-dimensional RBF64 dataset where it remains zero.

A quantitative summary is provided in Table \ref{tab:inc_sinc_combined_vertical}. To link these empirical results to our theory, we utilize the $n_{INC}$ metric---the tested ensemble size ($n \in \{2, 4, ..., 128\}$) closest to the theoretical \texttt{INC} value. The table reports the performance at this point as Accuracy @$n_{INC}$ (\% of Max), expressing the result as a percentage of the maximum accuracy achieved for that dataset. Finally, we report the Pearson correlation between the Probability of Linear Independence (PLI) and the average accuracy across all tested ensemble sizes ($n=2$ to $128$), providing a statistical measure of the relationship between diversity and performance.

\subsection{Discussion}
In this section, we interpret these empirical findings to evaluate the validity and practical utility of our theoretical framework. We first examine the fundamental correlation between algebraic diversity (PLI) and accuracy. We then assess the effectiveness of \texttt{INC} as a sizing heuristic for both robust (OzaBagging) and complex (GOOWE) ensemble methods, discuss the implications of high-dimensional spaces, and finally refine the interpretation of our theoretical estimators.

\vspace{5pt}
\subsubsection{\textit{Confirmation of the PLI-Accuracy Relationship}}

As a foundational check, we first confirm the relationship between PLI and accuracy. Across nearly all datasets and for both ensemble methods, a clear positive correlation is observed in the figures. As $n$ increases, the PLI (red curve) rises, empirically validating Theorem 3. As expected, the PLI is 0 for all $n < m$, as it is mathematically impossible to obtain $m$ linearly independent vectors from fewer than $m$ classifiers. This increase in PLI is mirrored by a rise in classification accuracy (blue curve).

This visual trend is quantitatively supported by the generally high Pearson correlation coefficients listed in Table \ref{tab:inc_sinc_combined_vertical}. Excluding the RBF64 dataset where PLI is universally 0, we observe a strong positive correlation ($>0.6$) in 17 out of the 22 experimental cases. This confirms the strong statistical link between achieving linear independence and improving accuracy, motivating our central hypothesis: that the saturation of this algebraic diversity should correspond to the saturation of performance.

The few exceptions where correlation is low (e.g., Poker with GOOWE, 0.034) are notably instructive. In these cases, the low correlation is not due to a lack of diversity, but rather the behavior of the weighting mechanism. As seen in Figure \ref{fig:goowe}, for Poker, the accuracy actually degrades at larger ensemble sizes even as the PLI continues to rise toward 1.0. This observation aligns with the established principle in ensemble theory that diversity is a necessary but not sufficient condition for performance. While high PLI ensures the potential to represent complex decision boundaries, it does not guarantee accuracy if the combination mechanism (in this case, GOOWE's weighting) fails to effectively aggregate that diversity.

\vspace{5pt}

\subsubsection{\textit{OzaBagging/Validating INC as a Practical Performance Heuristic}}
Table \ref{tab:inc_sinc_combined_vertical} provides a direct validation of \texttt{INC} as a practical heuristic. For OzaBagging, the results are compelling. In nearly all cases, the ensemble size closest to our theoretical \texttt{INC} ($n_{INC}$) achieves over 99\% of the maximum possible accuracy (and 100\% in some cases).

\begin{itemize}
    \item For RBF16 and RBF32, targeting the \texttt{INC} (34 and 66, respectively) leads to $n_{INC}$ values (32 and 64) that achieve 100.00\% of the maximum accuracy.
    \item For Click Prediction ($m=2$), our $n_{INC}=4$ achieves 99.91\% of the maximum.
    \item For Covtype ($m=7$), our $n_{INC}=32$ achieves 99.72\% of the maximum.
\end{itemize}

This provides powerful evidence that \texttt{INC} is a highly effective and practical heuristic for robust, majority-voting ensembles. It reliably identifies the point of full performance saturation, confirming that achieving a high degree of theoretical linear independence is a direct proxy for achieving peak empirical performance.

\vspace{5pt}

\subsubsection{\textit{GOOWE/The Instability of Complex Weighting}}
In contrast, the results for GOOWE reveal that achieving representational capacity is not always sufficient for optimal performance when using complex weighting schemes. While GOOWE shows high performance on some datasets (e.g., Covtype at 99.73\%, RBF32 at 100\%), it often fails to saturate at the theoretical diversity point identified by \texttt{INC}.

\begin{itemize}
    \item On RBF2 and RBF4, targeting \texttt{INC} (5 and 7) yields only 87.97\% and 89.01\% of the maximum accuracy.
    \item Similarly, on Airlines and Rialto, the performance at $n_{INC}$ is roughly 95\% of the maximum, with peak accuracy occurring at larger ensemble sizes (e.g., $n=64$).
\end{itemize}

This highlights a critical distinction. While \texttt{INC} guarantees that the ensemble \textit{can} algebraically represent the solution (necessary condition), GOOWE's geometric optimization may require additional factors to realize this potential. As Figure \ref{fig:goowe} shows, GOOWE's accuracy often peaks and then \textit{degrades} at larger ensemble sizes (e.g., Poker, RBF32). In some cases (e.g., Airlines), the method benefits from redundancy, requiring ensemble sizes larger than \texttt{INC} to stabilize the weights. In other cases (e.g., Poker, RBF32), the method succumbs to instability, where performance degrades as the ensemble grows. Unlike OzaBagging, GOOWE's performance is dependent on complex dynamics that decouple the point of peak accuracy from the point of algebraic saturation.

\vspace{5pt}

\subsubsection{\textit{The Challenge of High-Dimensionality (RBF64)}}
The RBF64 dataset ($m=64$) provides an edge-case. For both methods, the PLI (red curve) effectively remains 0 for all tested sizes, as spanning a 64-dimensional space is exceptionally difficult. Consequently, \texttt{INC} and \texttt{SINC} are incalculable.

However, despite this complete \textit{lack} of full linear independence, the ensembles still achieve high practical performance (as seen in Figure \ref{fig:ozabag} and \ref{fig:goowe}, with OzaBagging achieving over 96\% accuracy). This offers a crucial nuance to our theory: while Theorem 1 and Section III establish that $m$ independent votes are necessary for universal representational capacity (the ability to classify \textit{any} possible instance), the RBF64 results show that this full capacity is not always required for high practical accuracy. 

\vspace{5pt}

\subsubsection{\textit{Re-evaluating INC and SINC as Estimators}}
This analysis leads to a more nuanced interpretation of our theoretical metrics, \texttt{INC} and \texttt{SINC}. They should be understood as a theoretical benchmark for full representational capacity. The key takeaway is twofold:

\begin{enumerate}
    \item For robust methods like OzaBagging, our framework is validated: \texttt{INC} serves as a reliable, conservative target that guarantees full performance saturation (over 99\% of max accuracy).
    \item For complex methods like GOOWE, \texttt{INC} identifies the point of algebraic sufficiency, but not necessarily algorithmic optimality. Our results show that complex weighting schemes may require sizes larger than \texttt{INC} or smaller than \texttt{INC} to achieve their peak.
\end{enumerate}

This analysis also clarifies the role of \texttt{SINC}. As shown in Table \ref{tab:inc_sinc_combined_vertical}, \texttt{SINC} consistently provides a smaller, more optimistic estimate for the point of saturation compared to \texttt{INC} (i.e., $\texttt{SINC} \le \texttt{INC}$). This is a valuable practical feature, as \texttt{SINC} is derived from a simplified, closed-form approximation that is computationally far more efficient than the full \texttt{INC} model. 

We also note that due to the discrete nature of our tested ensemble sizes (e.g., 16, 32, 64), targeting the \texttt{SINC} value would have resulted in the same $n_{INC}$ values in our experiments. Therefore, \texttt{SINC} can be effectively used as a computationally cheaper lower-bound heuristic for \texttt{INC}.

\vspace{5pt}

Our framework is thus successful not just in predicting performance, but in establishing a theoretical landmark that helps explain and quantify the trade-offs between diversity, complexity, and performance saturation.

\section{CONCLUSION AND FUTURE WORK}
This paper introduced a theoretical framework that explains the trade-off between ensemble size and performance by modeling the linear independence of classifier vote vectors. We established that achieving a set of \textit{m} linearly independent votes is a necessary condition for an ensemble's representational capacity and developed a probabilistic model to determine the ensemble size required to meet this condition with a specified confidence level. 

Our empirical results confirmed a strong correlation between the Probability of Linear Independence (PLI) and accuracy, validating that our method can effectively identify the point of diminishing returns for majority voting ensembles and reveal limitations in more complex weighting schemes. Ultimately, this work provides a principled methodology for ensemble sizing that moves beyond simple heuristics.

A key simplification in our model is the assumption that the dependency probabilities, $p_l$, are uniform across all classifiers. This "homogeneity" assumption implies that classifiers are, on average, indistinguishable from one another. This first-order approximation allowed us to establish a clean and effective theoretical framework, which, as our experiments show, successfully models the behavior of robust ensembles like OzaBagging.

For the future work, this framework could be extended to a "heterogeneous" model. Investigating individual classifier-specific dependency probabilities ($p_{i,l}$) could provide a more detailed and accurate theoretical framework, potentially explaining the performance nuances of individual classifiers or more complex, non-linear weighting schemes.

\balance
\bibliographystyle{IEEEtran}
\bibliography{sample-base}

\end{document}